\documentclass[letterpaper, 10 pt, conference]{ieeeconf}  

\IEEEoverridecommandlockouts                              

\usepackage{epsfig} 
\usepackage{mathptmx} 
\usepackage{times} 
\usepackage{amsmath} 
\usepackage{amssymb}  
\usepackage{graphicx}
\usepackage{array}
\usepackage{booktabs}
\usepackage{longtable}
\usepackage{rotating}
\usepackage{multirow}    
\usepackage[table,xcdraw]{xcolor} 
\usepackage{orcidlink}
\usepackage{makecell}
\usepackage{xcolor}

\hypersetup{hidelinks,
	colorlinks=true,
	allcolors=black,
	pdfstartview=Fit,
	breaklinks=true
}
    
\title{\LARGE \bf
OpenVLN: Open-world Aerial Vision-Language Navigation
}
\author{
    Peican Lin$^{1}$, Gan Sun$^{1,*}$, Chenxi Liu$^{2}$, Fazeng Li$^{1}$, Weihong Ren$^{3}$ and Yang Cong$^{1}$ 
   \thanks{ $^{1}$School of Automation Science and Engineering, South China University of Technology,  {\tt\small \{paxon.lam, lifazeng818\}@gmail.com}}
    \thanks{$^{2}$State Key Laboratory of Robotics, Shenyang Institute of Automation, Chinese Academy of Sciences, {\tt\small liuchenxi0101@gmail.com}.}
    \thanks{$^{3}$State Key Laboratory of Robotics and System, School of Mechanical Engineering and Automation, Harbin Institute of Technology, Shenzhen.}
    \thanks{
    *Corresponding author: {\it Prof. Gan Sun}, {\tt\small sungan1412@gmail.com}.}
}

\begin{document}

\maketitle

\pagestyle{empty}

\begin{abstract}

Vision-language models (VLMs) have been widely-applied in ground-based vision-language navigation (VLN). However, the vast complexity of outdoor aerial environments compounds data acquisition challenges and imposes long-horizon trajectory planning requirements on Unmanned Aerial Vehicles (UAVs), introducing novel complexities for aerial VLN. To address these challenges, we propose a data-efficient \underline{Open}-world aerial \underline{V}ision-\underline{L}anguage \underline{N}avigation (\emph{i.e.,} OpenVLN) framework, which could execute language-guided flight with limited data constraints and enhance long-horizon trajectory planning capabilities in complex aerial environments. Specifically, we reconfigure a reinforcement learning framework to optimize the VLM for UAV navigation tasks, which can efficiently fine-tune VLM by using rule-based policies under limited training data. Concurrently, we introduce a long-horizon planner for trajectory synthesis that dynamically generates precise UAV actions via value-based rewards. To the end, we conduct sufficient navigation experiments on the TravelUAV benchmark with dataset scaling across diverse reward settings. Our method demonstrates consistent performance gains of up to 4.34\% in Success Rate, 6.19\% in Oracle Success Rate, and 4.07\% in Success weighted by Path Length over baseline methods, validating its deployment efficacy for long-horizon UAV navigation in complex aerial environments.


\end{abstract}

\section{INTRODUCTION}
Vision-language navigation (VLN)\cite{andersonVision2018} is a cornerstone task for embodied agents, it demands that agents traverse intricate, real-world environments solely via following natural-language instructions. Recent VLN research  has grew significantly, with notable proliferation of novel methodologies in this domain. A central focus of VLN research involves constructing world models that enable agents to comprehend their surrounding environments\cite{gaoOpenFly2025}. Simultaneously, the emergence of large pre-trained vision-language models (VLMs) has spurred a rise of VLN methodologies utilizing fune-tuning techniques\cite{LLaMAVIDImageWorth}. As the performance of pre-trained VLMs improves, researchers aim to develop agents capable of reasoning and planning trajectories from human instructions. 

\begin{figure}[t]
        \centering
        \includegraphics[width=3.4in]{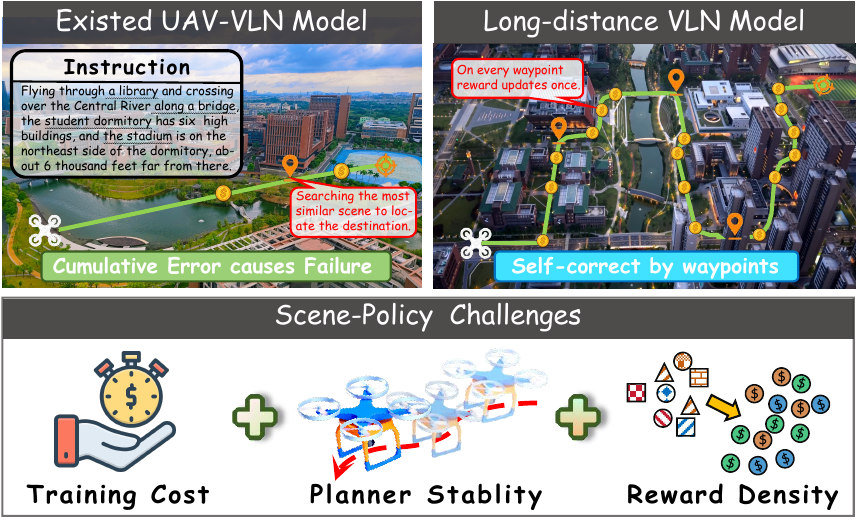}
        \caption{Challenge illustration for long-distance Vision-Language Navigation task, where drone needs to navigate to the destination according to the given instruction. Extended planning trajectories propagate environmental uncertainty, compounding localization error accumulation.} 
        \label{fig:motivation}
\end{figure}

Although current research has achieved substantial progress in grounded navigation\cite{zhuTargetdrivenVisualNavigation2017,NavilaLeggedRobot}, aerial navigation\cite{zhouEGOswarmFullyAutonomous2021} for unmanned aerial vehicles (UAVs) remains critically underexplored. Compared to ground-based VLN tasks, UAV-based VLN confronts significantly more complex environments. Compounding this challenge, practical aerial data collection difficulties result in \textbf{scarcity of adequate UAV-VLN datasets} that impede real-world implementation. Furthermore, conventional approaches \textbf{fail to maintain a precise planning trajectory during long-horizon navigation}, especially in unstructured aerial domains. While AerialVLN\cite{liuAerialVLNVisionandlanguageNavigation2023} and OpenUAV\cite{wangRealisticUAVVisionlanguage2024} establish vision-language navigation benchmarks for UAVs, their methodologies necessitate fine-tuning on large-scale simulated datasets. This constraint is unfavorable for real-world deployment due to the inherent challenges of aerial data acquisition\cite{wangUAVflowColosseoRealworld2025}. Furthermore, these approaches demonstrate limited long-horizon planning capabilities, resulting in pronounced performance degradation when confronted with extended navigation tasks.

To further analyze these practical challenges, we investigate aerial VLN for UAVs under dual constraints of \textbf{data scarcity} and \textbf{long-horizon planning requirements}. 
Specifically: (1) Data scarcity impedes cross-modal representation learning, degrading the accuracy of navigation task under sparse training dataset; (2) Extended planning trajectory propagate environmental uncertainty, compounding localization error accumulation across extended trajectories. Conventional approaches\cite{guptaCognitiveMappingPlanning2017,dorbala2022clipnavusingclipzeroshot,yueSafeVLNCollisionAvoidance2024} fail to jointly address these issues as supervised VLM adaptation lacks policy generalization and classical planners face limitations with mission-scale contingencies.

\begin{figure*}[ht]
\centering
\includegraphics[width=17.8cm]{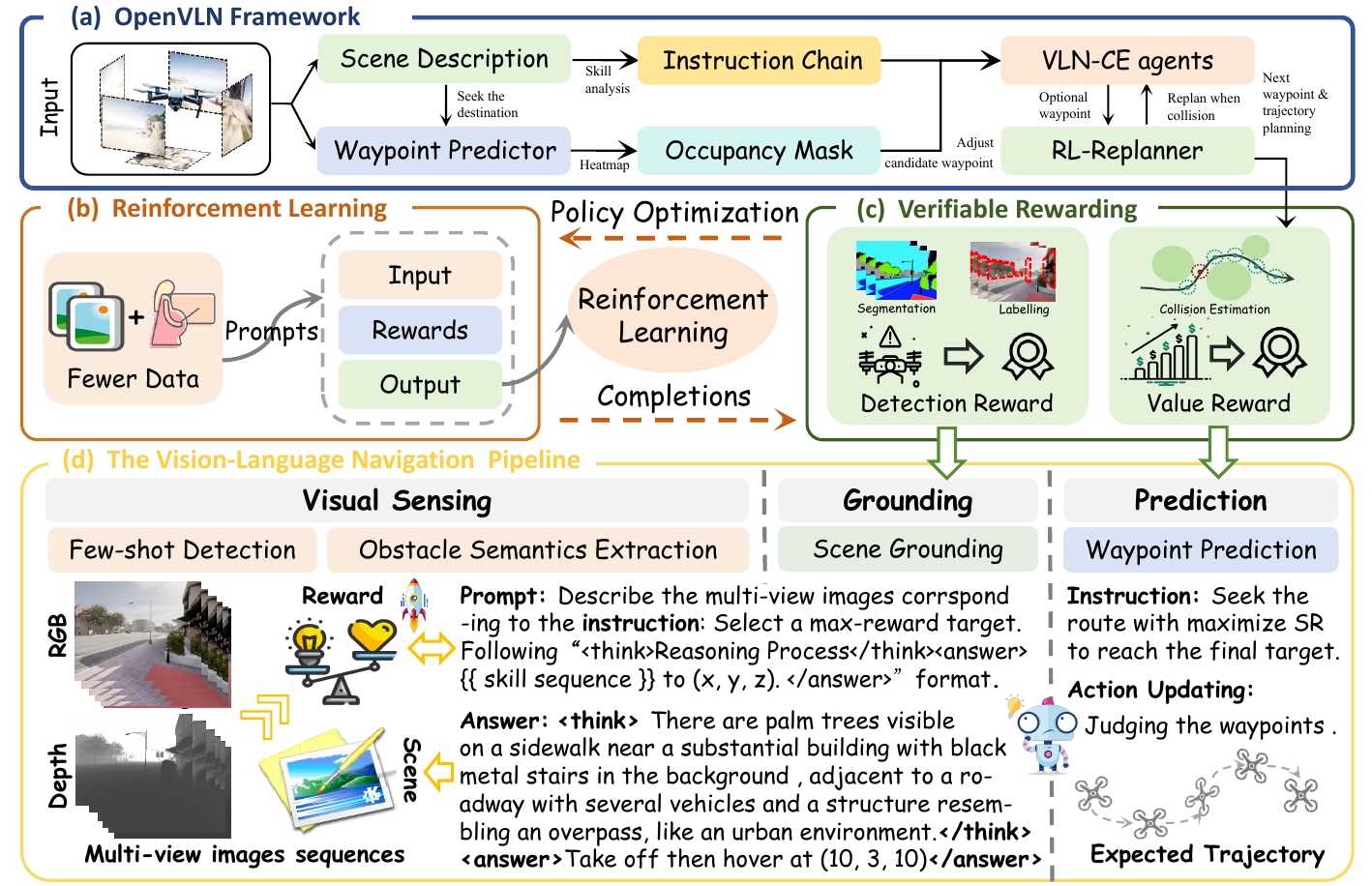}
\caption{\textbf{Overall architecture of our OpenVLN framework.} (a) The VLN-CE replanner that controls flight actions; (b) and (c) together form an Reinforcement Learning framework for data-efficient VLM fine-tuning under data scarcity, including a verifiable reward, value model reward at waypoints and an RL fine-tuning optimizer; and (d) a vision-language navigation model with a sensing encoder, multi-modal grounding module\cite{liuGroundingDINOMarrying2025}, and an action decoder that predicts the next waypoint, producing the planned trajectory and the UAV pose. }
\label{fig:framework}
\end{figure*}

To resolve the challenges above, in this paper, we develop an \underline{Open}-world aerial \underline{V}ision-\underline{L}anguage \underline{N}avigation (OpenVLN) framework. Our framework introduces a novel approach to the OpenUAV benchmark, leveraging reinforcement learning and a value reward mechanism to enable efficient VLM fine-tuning under data scarcity constraints. On the one hand, to efficiently fine-tune the VLM, we design a rule-based reinforcement learning method that optimizes VLM updates through a novel reward function. Departing from conventional reward mechanisms, this approach streamlines policy construction while enabling self-evolution during model fine-tuning, extracting robust task-aligned feature representations from minimal data samples. This approach substantially outperforms conventional supervised learning methods, demonstrating enhanced robustness when only sparse training data are available. On the other hand, we tackle long-horizon navigation challenges by introducing a value-based reward optimizer that synthesizes extended trajectory sequences. Unlike conventional VLN methods that supervise trajectories only at the final waypoint, our approach provides dense, per-waypoint supervision: every waypoint along the trajectory is evaluated by a learned value model, yielding verifiable intermediate rewards, better credit assignment over long horizons, and improved stability. The value optimizer evaluates waypoint quality and provides dense feedback signals by fusing visual observations with instruction features during planning. A value model is utilized to assess state quality and guide optimal path selection under environmental uncertainty in long-range trajectory planning. This enhances the accuracy of UAV trajectory planning, enabling the generation of smoother, more robust trajectories. Consequently, it facilitates efficient path planning for complex navigation tasks during UAV operations. The main contributions of this paper are as follows:
\begin{itemize}
\item We propose an \underline{Open}-world aerial \underline{V}ision-\underline{l}anguage \underline{N}avigation (\emph{i.e.,} OpenVLN) framework, a novel architecture that synergizes reinforcement learning for efficient VLM fine-tuning with long-horizon trajectory VLN-CE planning for UAV navigation.
\item  We develop a self-evolving reinforcement learning optimization strategy with task-aligned rewards, enabling robust VLM fine-tuning using minimal aerial data while outperforming supervised methods. 
\item We introduce a value-model–guided long-horizon reward mechanism that combines dense per-waypoint reward-estimates with a verifiable reward, guiding optimal waypoint selection and improving accuracy and stability in long-horizon UAV navigation.
\end{itemize}



\section{RELATED WORK}

\subsection{Vision-Language Navigation}

Researchers commonly employ pretrained vision-language models (VLMs) and fine-tune them for vision-language navigation (VLN), yet real-world UAV scenarios suffer from \textbf{severe data scarcity}. Early VLN datasets, such as R2R \cite{andersonVision2018}, REVERIE \cite{qiREVERIERemoteEmbodied2020}, and SOON \cite{zhuSOONScenarioOriented2021a}, primarily focused on indoor environments, equipping agents with complex scene exploration abilities following natural language instructions.
In , 
AerialVLN\cite{liuAerialVLNVisionandlanguageNavigation2023} first brings discrete-action, long-trajectory navigation to aerial scenes; CityNav\cite{lee2025citynavlargescaledatasetrealworld} scales it to real-world maps yet still relies on fixed action sets and pre-computed global plans.
However, their reliance on fixed action sets and precomputed long-range plans limits adaptability to diverse outdoor distributions.
CMA-based models\cite{dorbala2022clipnavusingclipzeroshot} require large, diverse paired image-instruction examples to learn stable alignments but often overfit or forget rare visual cues in limited UAV datasets. End-to-end policies \cite{saxena2025uavvlnendtoendvisionlanguage}, which map raw pixels and language directly to low-level controls, demand heavy supervision and extensive training, leading to slow convergence and brittleness to imperfect localization and distribution shifts .
To address these limitations, recent work has transitioned from discrete nav-graphs to continuous settings, exposing agents to low-level control and richer distributional shifts \cite{krantz2020, krantzSim2simTransferVisionandlanguage2022}. Many methods have adopted a natural hierarchical remedy, including Zhou et al. \cite{zhouNavGPTExplicitReasoning2024}, Wang et al. \cite{wangRealisticUAVVisionlanguage2024, wangUAVflowColosseoRealworld2025}, and Zhang et al. \cite{zhangNaVidVideobasedVLM2024}, who trained a waypoint predictor module to obtain local candidate waypoints. Conventionally, the VLN task has usually been formulated as a supervised learning problem, forcing the agent independent from the map visual observations and language instructions to actions.

\subsection{Reinforcement Learning}
Reinforcement learning (RL) has emerged as a pivotal approach in visual learning and embodied navigation, particularly in addressing complex tasks that \textbf{require the optimization of long-term objectives through interaction} \cite{andersonVision2018}. However, the inherent challenges of RL, including sparse or delayed rewards, long-horizon credit assignment, and the need for extensive exploration, result in low sample efficiency and high computational costs\cite{liuRDT1BDiffusionFoundation2025}. These issues are further exacerbated in UAV visual language navigation (VLN), where distribution shifts and partial observability complicate the learning process \cite{kulkarniDeepSuccessorReinforcement2016}. 
Recent advances in DeepSeek-R1 model\cite{deepseek-aiDeepSeekR1IncentivizingReasoning2025} have explored RL-first training paradigms that bypass traditional supervised fine-tuning, thereby reducing computational overhead and accelerating the training process \cite{chenConRFTReinforcedFinetuning2025}. These approaches leverage consistency and KL regularization, along with externally verifiable signals, to implement stable policy updates. Drawing inspiration from this methodology, our work adopts RL as the primary mechanism for adaptation, aiming to achieve efficient and controlled policy improvement within limited interaction budgets. 
In the realm of reward construction for VLN, two complementary approaches have been developed. Value-model-based methods, such as those proposed by \cite{kulkarniDeepSuccessorReinforcement2016}, employ value or successor-feature estimators to convert sparse outcomes into dense feedback. This transformation facilitates improved exploration and credit assignment over long horizons. Similarly, \cite{ichterCanNotSay2023} extended this concept through potential shaping and value-based feedback mechanisms. 
In parallel, loss-based reward formulations have been instrumental in the derivation of dense, verifiable signals from training objectives. These methods, as demonstrated in \cite{zangInternLMXComposer25rewardSimpleEffective2025}, utilize contrastive consistency, distance-to-go reductions, and KL-regularized updates to guide stable RL fine-tuning while minimizing reliance on extensive human annotations. 
Our approach integrates both methodologies: a value model provides dense\cite{shuRFTFReinforcementFinetuning2025}, verifiable rewards\cite{liuVisualRFTVisualReinforcement2025} to evaluate the quality of the waypoint, while a KL-regularized RL objective\cite{ouyang2022training} ensures stable policy updates under data constraints. Additionally, we incorporate a VLN-CE planner to model smooth trajectory distributions, addressing the issue of compounding errors in long-horizon planning. This comprehensive strategy yields reliable waypoint synthesis, thereby advancing UAV VLN capabilities. 
\begin{figure*}[h]
        \centering
        \includegraphics[width=17.8cm]{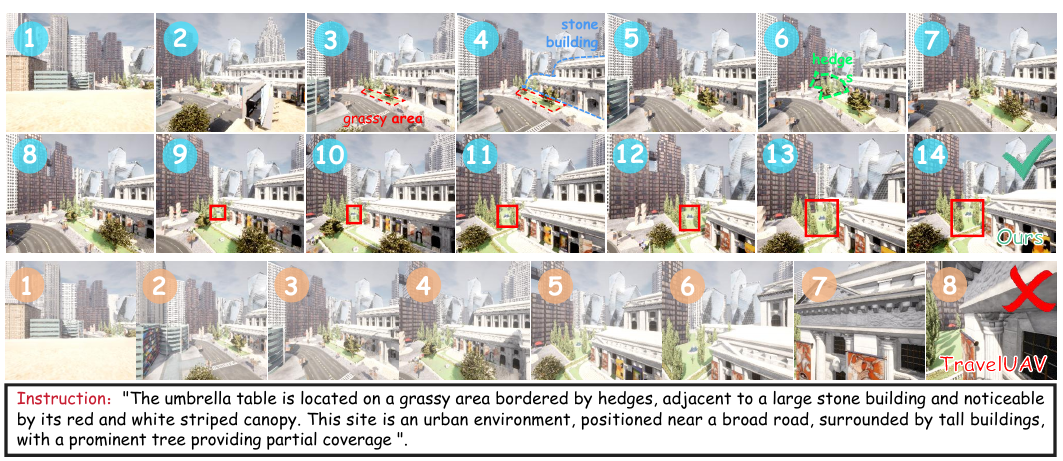}
        \caption{\textbf{Comparison between the baseline and our proposed method.} Rows 1–2: our UAV navigator successively searches and reliably reaches the destination by progressively detecting the instructed objectives one by one.
        Row 3: with the baseline method the drone collides with the building; the mission fails because of overlapping occlusion.}
 \end{figure*}
\section{METHODOLOGY}
\subsection{Overview}
In this paper, we focus on the UAV vision-language navigation (VLN) task. This task requires a UAV to navigate through a complex environment guided by natural language instructions. 
To tackle the challenges of data scarcity and long-horizon planning of UAV, we propose the Open-world aerial Vision-Language Navigation framework. Our approach integrates reinforcement learning for efficient vision-language model (VLM) fine-tuning under fewer data conditions, alongside a value model based long-horizon trajectory planning. The overall architecture of the OpenVLN framework is illustrated in Fig. \ref{fig:framework}.
We design a value model to construct a dense reward function addressing the challenge of sparse rewards in long-horizon navigation tasks.  This model evaluates the current state quality of each waypoint, providing more informative feedback to guide the navigation decisions.
The RL fine-tuning utilizes a rule-based policy to optimize VLM updates through a verified reward function, enabling self-evolution during model fine-tuning. 
The proposed method improves the accuracy of trajectory planning, enabling the generation of smoother and robust trajectories. Consequently, it facilitates efficient path planning for complex navigation tasks.

\subsection{Preliminary} 
\noindent \textbf{Problem Formulation.} \; The objective of VLN is to enable the UAV to interpret and execute these instructions effectively, reaching specified waypoints or destinations while avoiding obstacles and adhering to environmental constraints. Formally, the UAV VLN task can be defined as follows: Given a natural language instruction $I$ and an initial state $s_0$ of the UAV, the goal is to learn a policy $\pi_\theta(a_t|s_t, I)$ that maps the current state $s_t$ and instruction $I$ to an action $a_t$ at each time step $t$.
The state $s_t$ typically includes the position and orientation of UAV, while the action $a_t$ represents the control commands that dictate the UAV's movement. State $s_t$ and action $a_t$ share the same space $\mathcal{S}=\{x,y,z,\theta,\phi,\psi\}$, $x,y,z$ represents the 3D coordinates of the UAV in the environment, while $\theta,\phi,\psi$ denote the Euler angles representing the orientation of UAV.
The navigation process continues until the UAV reaches the goal or a maximum number of steps $T$ is reached. The performance of the policy $\pi_\theta$ is evaluated based on the success rate of reaching the goal and the efficiency of the trajectory taken.

\noindent \textbf{Reinforcement Learning with Verifiable Rewards.} \;  Reinforcement Learning with Verifiable Rewards (RLVR) is a novel reinforcement learning framework that incorporates verifiable rewards to enhance the learning process of agents in complex environments. 
The key idea behind RLVR is to design a reward function that can be easily verified by an external oracle or expert, ensuring that the agent receives accurate feedback on its actions. This approach addresses the challenge of sparse and delayed rewards commonly encountered in traditional reinforcement learning settings, where agents may struggle to learn effective policies due to limited feedback. In RLVR, the reward function is constructed based on a set of verifiable criteria that reflect the desired behavior of the agent. By incorporating verifiable rewards, RLVR enables agents to receive more informative feedback, allowing them to learn more efficiently and effectively.

\noindent \textbf{UAV Navigation Simulator.} \;  UAV Navigation Simulator (UNS) is a comprehensive simulation platform designed for testing and evaluating the planning performance in various navigation scenarios. Among them, AirSim-based UAV simulators provide photorealistic rendering, high-fidelity physics, and controllable environments to evaluate navigation across diverse scenes, sensors, and conditions. Recent variants integrate VLN, enabling agents to follow natural language in complex environments. We use the TravelUAV simulator for all experiments to assess OpenVLN under realistic, data-scarce, and long-horizon settings.

\subsection{Value Model based Long-Horizon Trajectory Planning} 
A remaining problem is how to evaluate the rationality of waypoint selection and its impact on subsequent path planning during UAV navigation to the destination.
To address this issue, we introduce a value model that evaluates the quality of the current waypoint, guiding the UAV to select the optimal path and thereby enhancing the accuracy of long-horizon navigation.

\begin{figure*}[!t]
        \centering
         \includegraphics[width=17.8cm]{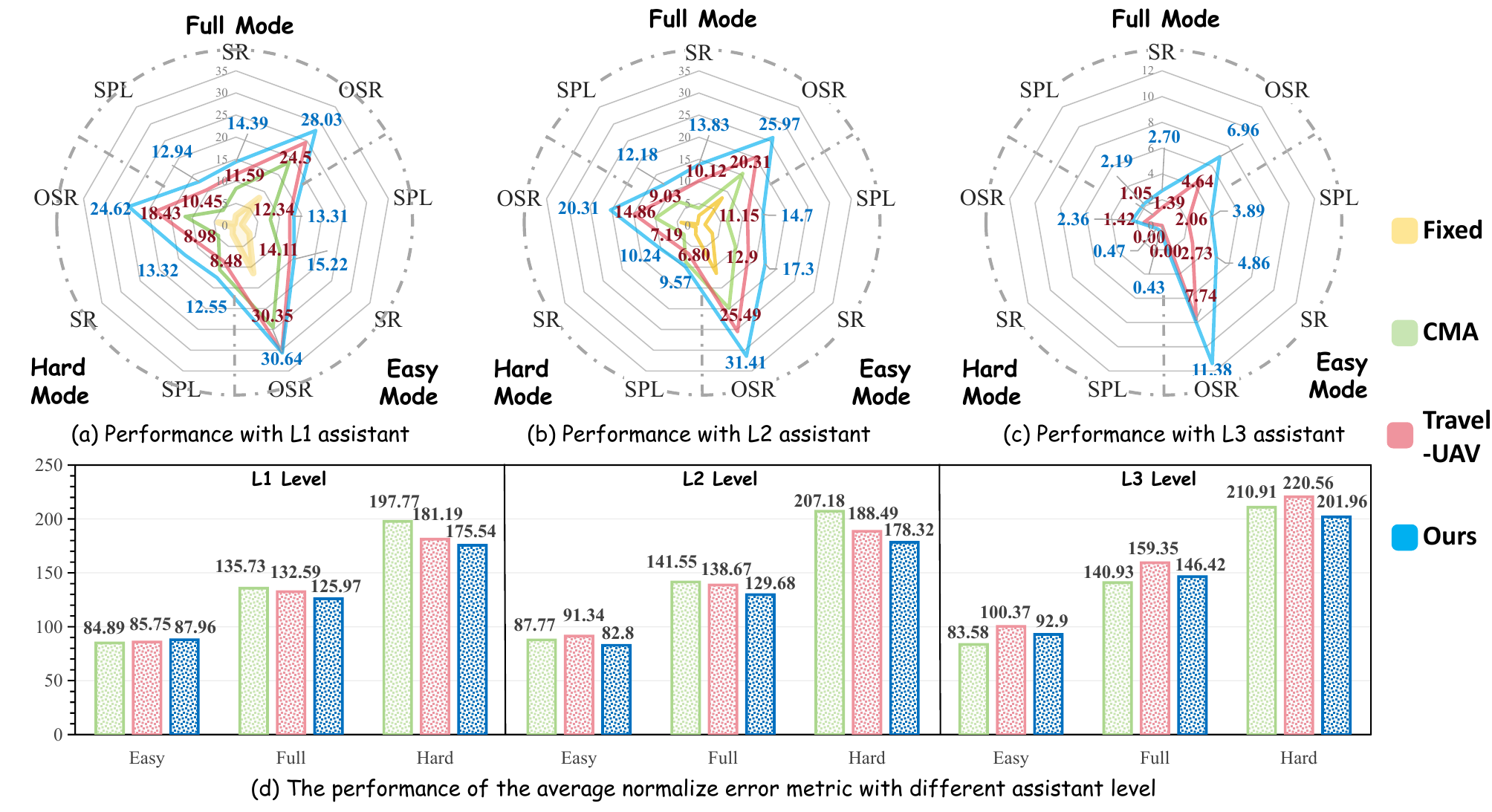}
\caption{\textbf{The evaluation results with three levels of the assitance respectively.} (a)-(c) The radar diagram demonstrating the Success Rate (SR), Oracle Success Rate (OSR) and Success weighted by Path Length (SPL) performances with Level 1-3 assitance, higher is better. (d) The Average Normalized Error in the TravelUAV benchmark evaluation, lower is better.} 
\label{fig:eval}
\end{figure*}
To convert sparse rewards from outcomes into dense rewards, we design a value model $V_\rho$ that evaluates the value of the current state. Specifically, given the current state $s_t$ and encoded multimodal input $T_\text{input}=<T_\text{img},T_{I}>$, the value model outputs the estimated value of the state as $v_t=V_\rho(s_t,T_\text{input})$.
During long-range flights, solely relying on terminal state rewards may lead to ineffective evaluation of intermediate states, thereby impacting the accuracy of path planning. To address this issue, we assume that within a trajectory, states closer to the terminal state should possess higher values. As time progresses, the UAV should move closer to the terminal state to complete the task; thus, values should monotonically increase over time, implying that each decision moves closer to the terminal state, i.e., $v_t<v_{t+1}<\cdots<v_{t+n-1}$. The value of a state depends not only on its current waypoint value but also on the cumulative value of all reachable waypoints from that point onward. Based on this assumption, we apply a reward function for long-horizon navigation:
\begin{equation}
        R_{\text{t}}^{V_{\rho}} = \sum_{t=1}^{n} \gamma^{t} \cdot \left( V_\rho(s_t) - V_\rho(s_{t+1}) \right),
\end{equation}
where $\gamma$ is a discount factor that balances immediate and future rewards, and $n$ is the length of the trajectory. We apply the reward function on every waypoint cumulatively, which encourages the UAV to select waypoints that lead to higher overall value.
By minimizing this loss function, we encourage the value model to assign higher values to states closer to the terminal state, thereby improving its ability to evaluate the quality of waypoints during long-horizon trajectory planning. Since the value function considers all future rewards with discounting, it guides the UAV to select a path that maximizes cumulative value rather than merely choosing the next waypoint with the highest immediate value. This approach addresses the issue of local optima, ensuring global optimality in long-horizon navigation.

\subsection{Reinforcement Fine-tuning with Verifiable Rewards}
Due to the scarcity of real-world data, traditional supervised learning methods struggle to effectively fine-tune policy models. To address this issue, we introduce a reinforcement learning pipeline that employs a verifiable reward function, optimizing the model under limited data conditions and thereby enhance its performance in navigation tasks.

Based on the value assessments provided by the value model for each state, we design a verifiable reward function to construct dense rewards.
Specifically, given the current state $s_t$ and multimodal input $T_\text{input}$, the value model outputs the estimated value of the state as $v_t$. 
The state $s_t$ and the target waypoint $w_n$ are input into a pre-trained multimodal large language model (MLLM) to obtain their feature representations $F_{s_t}$ and $F_{w_n}$:
\begin{equation}
        \begin{aligned}
        F_{s_t} &= \text{MLLM}(T_\text{input}, s_t),\\
        F_{w_n} &= \text{MLLM}(T_\text{label}, w_n), 
        \end{aligned}
\end{equation}
where $T_\text{label}$ is the encoded multimodal input corresponding to the ground-truth waypoint $w_n$. 
Then, we compute the cosine similarity between the two feature representations. A higher similarity between the two feature representations indicates a better alignment between the current state and the target waypoint, resulting in a larger reward. However, the reward magnitude should be constrained within a certain range to prevent excessively large rewards from adversely affecting the training process. A reward gain threshold is set to enable verifiable reinforcement learning fine-tuning. Specifically, we define the reward function as follows:
\begin{equation}
    \label{eq:dense-reward-machanism}
      R_t^{V} = \left\{
        \begin{aligned}
                r_\text{level}\qquad, \quad & \text{if } \frac{1}{1-\text{Sim}  (F_{s_t},F_{w_n})} \geq r_\text{level},  \\
                \frac{1}{1-\text{Sim}(F_{s_t},F_{w_n})}, \quad & \text{otherwise}.
        \end{aligned}
        \right.
\end{equation}

Following the RLVR framework, we define the optimization objective for reinforcement learning fine-tuning of the policy model as follows:
\begin{table*}[t]
\centering
\caption{\textbf{Results across Level 1-3 assistant.}
1) Random and Fixed method is basic method.
2) Cross-Modal Attention (CMA) model\cite{andersonVision2018} is commonly used in grounded VLN tasks and employs a bi-directional LSTM to simultaneously process image inputs and instruction comprehension.
3) TravelUAV \cite{wangRealisticUAVVisionlanguage2024} is a state-of-the-art method and provide an assistant-guided realistic UAV-VLN benchmark, utilizing a hierarchical framework with a waypoint predictor.
Performance improvements of OpenVLN are highlighted in cyan, below the metric value in bold, with arrows indicating the direction of change. }
\label{tab:4-1:Ours_eval_results}
\setlength{\tabcolsep}{2pt}
\begin{tabular}{cccccccccccccccc}
\toprule
\multirow{2}{*}{Method} & \multirow{2}{*}{Assistant} & \multicolumn{4}{c}{Full} & \multicolumn{4}{c}{Easy} & \multicolumn{4}{c}{Hard} \\
\cmidrule(lr){3-6} \cmidrule(lr){7-10} \cmidrule(lr){11-14}
& & NE $\downarrow$ & SR(\%)$\uparrow$ & OSR(\%)$\uparrow$ & SPL(\%)$\uparrow$ & NE $\downarrow$ & SR(\%)$\uparrow$ & OSR(\%)$\uparrow$ & SPL(\%)$\uparrow$ & NE $\downarrow$ & SR(\%)$\uparrow$ & OSR(\%)$\uparrow$ & SPL(\%)$\uparrow$ \\
\midrule
Random(100\%)      & L1 & 222.20  & 0.14  & 0.21  & 0.07  & 142.07 & 0.26  & 0.39  & 0.13  & 320.12 & 0.00   & 0.00     & 0.00 \\
Fixed(100\%)       & L1 & 188.61 & 2.27  & 8.16  & 1.40  & 121.36 & 3.48  & 11.48 & 2.14  & 270.69 & 0.79  & 4.09  & 0.49  \\
CMA (100\%)        & L1 & 135.73 & 8.37  & 18.72 & 7.90  & 84.89  & 11.48 & 24.52 & 10.68 & 197.77 & 4.57  & 11.65 & 4.51  \\
CMA (100\%)       & L2 & 141.55 & 7.02  & 15.39 & 6.54  & 87.77  & 9.55  & 19.87 & 8.74  & 207.18 & 3.94  & 9.92  & 3.94  \\
CMA (100\%)       & L3 & 140.93 & 4.89  & 11.56 & 4.41  & 83.58  & 7.35  & 17.81 & 6.53  & 210.91 & 1.89  & 3.94  & 1.83  \\
\midrule
TravelUAV (25\%)             & L1                                               & 132.59 & 11.59 & 24.50  & 10.45 & 85.75  & 14.11 & 30.35 & 12.34 & 181.19 & 8.98                     & 18.43                    & 8.48                     \\
TravelUAV (25\%)             & L2                                               & 138.67 & 10.12                     & 20.31 & 9.03  & 91.34  & 12.9  & 25.49 & 11.15 & 188.49 & 7.19                     & 14.86                    & 6.80                      \\
TravelUAV (25\%)             & L3                                               & 159.35 & 1.39                      & 4.64  & 1.05  & 100.37 & 2.73  & 7.74  & 2.06  & 220.56 & 0.00 & 1.42 & 0.00 \\ 
\midrule
Ours (25\%) & L1 &
\makecell{\bf 125.97\\\textcolor{cyan}{\bf ↓6.62}} &
\makecell{\bf 14.39\\\textcolor{cyan}{\bf ↑2.80}} &
\makecell{\bf 28.03\\\textcolor{cyan}{\bf ↑3.53}} &
\makecell{\bf 12.94\\\textcolor{cyan}{\bf ↑2.49}} &
\makecell{\bf 87.96\\\textcolor{magenta}{\bf ↑2.21}} &
\makecell{\bf 15.22\\\textcolor{cyan}{\bf ↑1.11}} &
\makecell{\bf 30.64\\\textcolor{cyan}{\bf ↑0.29}} &
\makecell{\bf 13.31\\\textcolor{cyan}{\bf ↑0.97}} &
\makecell{\bf 175.54\\\textcolor{cyan}{\bf ↓5.65}} &
\makecell{\bf 13.32\\\textcolor{cyan}{\bf ↑4.34}} &
\makecell{\bf 24.62\\\textcolor{cyan}{\bf ↑6.19}} &
\makecell{\bf 12.55\\\textcolor{cyan}{\bf ↑4.07}} \\

Ours (25\%) & L2 &
\makecell{\bf 129.68\\\textcolor{cyan}{\bf ↓8.99}} &
\makecell{\bf 13.83\\\textcolor{cyan}{\bf ↑3.71}} &
\makecell{\bf 25.97\\\textcolor{cyan}{\bf ↑5.66}} &
\makecell{\bf 12.18\\\textcolor{cyan}{\bf ↑3.15}} &
\makecell{\bf 82.80\\\textcolor{cyan}{\bf ↓8.54}} &
\makecell{\bf 17.30\\\textcolor{cyan}{\bf ↑4.40}} &
\makecell{\bf 31.41\\\textcolor{cyan}{\bf ↑5.92}} &
\makecell{\bf 14.70\\\textcolor{cyan}{\bf ↑3.55}} &
\makecell{\bf 178.32\\\textcolor{cyan}{\bf ↓10.17}} &
\makecell{\bf 10.24\\\textcolor{cyan}{\bf ↑3.05}} &
\makecell{\bf 20.31\\\textcolor{cyan}{\bf ↑5.45}} &
\makecell{\bf 9.57\\\textcolor{cyan}{\bf ↑2.77}} \\

Ours (25\%) & L3 &
\makecell{\bf 146.42\\\textcolor{cyan}{\bf ↓12.93}} &
\makecell{\bf 2.70\\\textcolor{cyan}{\bf ↑1.31}} &
\makecell{\bf 6.96\\\textcolor{cyan}{\bf ↑2.32}} &
\makecell{\bf 2.19\\\textcolor{cyan}{\bf ↑1.14}} &
\makecell{\bf 92.90\\\textcolor{cyan}{\bf ↓7.47}} &
\makecell{\bf 4.86\\\textcolor{cyan}{\bf ↑2.13}} &
\makecell{\bf 11.38\\\textcolor{cyan}{\bf ↑3.64}} &
\makecell{\bf 3.89\\\textcolor{cyan}{\bf ↑1.83}} &
\makecell{\bf 201.96\\\textcolor{cyan}{\bf ↓18.60}} &
\makecell{\bf 0.47\\\textcolor{cyan}{\bf ↑0.47}} &
\makecell{\bf 2.36\\\textcolor{cyan}{\bf ↑0.94}} &
\makecell{\bf 0.43\\\textcolor{cyan}{\bf ↑0.43}} \\      
\bottomrule
\end{tabular}
\end{table*}
\begin{equation}
\begin{aligned}
\label{eq:ppo-loss}
&-\mathbb{E}_{(s_t, a_t) \sim {P}} \Bigg\{ \min \left[ \frac{\pi_\theta(a_t|s_t)}{\pi_{\theta_{\text{old}}}(a_t|s_t)} R_t^{V}, \right. \nonumber \\
&\qquad\qquad\qquad \left. \text{clip} \left( \frac{\pi_\theta(a_t|s_t)}{\pi_{\theta_{\text{old}}}(a_t|s_t)}, 1 - \epsilon, 1 + \epsilon \right) R_t^{V} \right] \nonumber \\
&\qquad\qquad\qquad - \beta D_{\text{KL}} \left[ \pi_\theta(a_t|s_t) \| \pi_{\theta_{\text{ref}}}(a_t|s_t) \right] \Bigg\},
\end{aligned}
\end{equation}
where $\pi_{\text{ref}}$ is a reference policy model, and $\beta$ and $\epsilon $ are hyperparameters that balance the reward maximization and the KL divergence regularization, clip restricts the range of policy changes which can effectively avoid excessive updates and maintain the stability of the training process.

By optimizing this objective function, we enable the policy model to learn effective navigation strategies under limited data conditions, leveraging the dense rewards provided by the value model.
\section{Experiments}

\subsection{Experimental Setup}

\noindent \textbf{Benchmarks.} \; To ensure fair comparison under constrained data regimes, we re-implement and re-train the baseline method TravelUAV\cite{wangRealisticUAVVisionlanguage2024} on the same 25\% data partition, as the metric reference. Totally, we adopt four common benchmarks, include Random, Fixed, Cross-Modal Attention (CMA) model and TravelUAV (state-of-the-art).

\noindent \textbf{Dataset.} \; We directly adopt the TravelUAV benchmark and its UAV-Need-Help dataset as the evaluation corpus. The dataset contains 12,149 trajectories with diverse scene types and object annotations: the training set has 9,152 trajectories covering 76 objects in 20 scenes; the Test-Seen set has 1,410 trajectories drawn from the training set; the Test-Unseen-Map set has 958 trajectories from 2 unseen scenes; and the Test-Unseen-Object set contains 629 trajectories. \underline{Trajectories are split by length:} those under 250\,m are labeled as easy, and those at or above 250\,m as hard. Target distances range roughly from 50 to 400\,m. To simulate constrained-data conditions and validate data efficiency, we randomly sample only 25\% of trajectories from each scene as the training set (using a fixed seed for reproducibility); during evaluation, we fully adhere to the original benchmark’s data splits and test-set configurations. All experiments report results on both easy and hard subsets within each split. Metrics paragraphs follow in the main text.

\noindent \textbf{Training.} \; As an economical VLN solution, we exploit a memory-allocation mechanism tailored to small-scale GPU resources. The model is trained on 4$\times$ NVIDIA RTX 4090 GPU with 24GB memory, using the Adam optimizer with a learning rate of 1e-4 and a batch size of 1.  To verify the long-distance continual navigation ability, we set the maximum number of steps to 200 during training and evaluation. In the reinforcement learning fine-tuning stage, we set the KL regularization coefficient $\beta$ to 0.1. During training, we apply data augmentation techniques, including random cropping and horizontal flipping, to enhance model robustness. In the 25\% data scale setting, our value model training process consumes nearly 28 hours.

\noindent \textbf{Evaluation.} \; Evaluation is conducted on the Test Seen, Test Unseen Map, and Test Unseen Objects splits under three assistance levels that progressively remove external guidance: Level 1 supplies both the instruction and the groundtruth (GT) guidance ; Level 2 provides only the GT helper and Level 3 demands fully autonomous navigation, creating nine test conditions that jointly measure robustness and generalization. Limited by the UnrealEngine's port communication simplex mode, it can only be evaluated on a single NVIDIA 4090 GPU . 
\begin{table*}[t]
        \centering
        \caption{Results on ablation experiments with several reward function settings on the 25\% of dataset across Level 1-3 assistant.}
        \label{tab:4-2:ablation-results}
        \setlength{\tabcolsep}{2pt}
        \begin{tabular}{cccccccccccccccc}
        \toprule
        \multirow{2}{*}{Method} & \multirow{2}{*}{Assistant} & \multicolumn{4}{c}{Full} & \multicolumn{4}{c}{Easy} & \multicolumn{4}{c}{Hard} \\
        \cmidrule(lr){3-6} \cmidrule(lr){7-10} \cmidrule(lr){11-14}
        & & NE $\downarrow$ & SR(\%)$\uparrow$ & OSR(\%)$\uparrow$ & SPL(\%)$\uparrow$ & NE $\downarrow$ & SR(\%)$\uparrow$ & OSR(\%)$\uparrow$ & SPL(\%)$\uparrow$ & NE $\downarrow$ & SR(\%)$\uparrow$ & OSR(\%)$\uparrow$ & SPL(\%)$\uparrow$ \\
        \midrule
        5.0  & L1  & 125.97 & 14.39 & 28.03 & 12.94 & 87.96  & 15.22 & 30.64 & 13.31 & 175.54 & 13.32 & 24.62 & 12.55 \\
        5.0 & L2  & 129.68 & 13.83 & 25.97 & 12.18 & 82.80  & 17.30 & 31.41 & 14.70 & 178.32 & 10.24 & 20.31 & 9.57 \\
        5.0  & L3  & 146.42 & 2.70 & 6.96 & 2.19  & 92.90   & 4.86  & 11.38 & 3.89  & 201.96 & 1.47 & 3.36 & 1.43 \\ 
        \midrule
        3.0    & L1   & 158.35 & 6.85 & 14.26  & 6.12  & 103.95 & 8.67  & 17.59  & 7.26  & 214.80  & 6.04 & 11.47 & 5.21 \\
        3.0   & L2   & 155.49 & 3.39 & 10.12  & 2.89  & 102.05 & 6.79  & 16.68  & 5.52  & 210.95 & 6.36 & 9.16 & 5.68 \\
        3.0  & L3    & 174.13 & 1.93 & 3.74  & 1.04 & 112.33 & 3.42  & 6.12  & 3.18  & 238.27 & 1.08 & 3.88 & 0.75 \\
        \midrule
        1.0 & L1   & 184.33 & 4.23 & 9.39  & 4.98  & 122.76 & 5.67  & 10.73  & 4.65  & 248.61 & 4.18 & 6.54 & 4.02 \\
        1.0 & L2  & 188.05 & 2.55 & 7.23  & 2.13  & 121.49 & 2.20   & 4.46  & 4.25  & 257.31 & 2.36 & 3.08 & 2.04 \\
        1.0    & L3  & 189.10  & 1.55 & 7.08  & 2.14  & 128.40  & 2.20   & 3.16  & 2.78  & 258.59 & 0.86 & 1.64 & 0.86 \\ 
        \midrule
        $\infty$ & L1  & 197.08 & 1.31 & 2.54  & 1.05  & 129.31 & 3.61  & 4.12  & 3.30  & 267.41 & 1.18 & 1.47 & 0.89 \\
        $\infty$ & L2 & 199.46 & 0.90 & 1.55 & 0.61 & 132.47 & 1.37 & 2.88 & 1.20 & 273.13 & 0.32 & 0.76 & 0.66 \\
        $\infty$ & L3 & 200.97 & 0.73 & 1.24 & 0.84 & 133.83 & 1.82 & 2.28 & 1.65 & 275.83 & 0.08 & 0.16 & 0.04 \\
        \bottomrule
        \end{tabular}
\end{table*}

\noindent \textbf{Metrics.} \; To comprehensively assess model performance, we employ several widely used metrics \cite{andersonVision2018,krantz2020} : (1) Success Rate (SR) is the percentage of successful navigation episodes; (2) Oracle Success Rate (OSR): the success rate considering the closest point to the target reached during navigation; (3) Success weighted by Path Length (SPL) is a measure that accounts for both success and efficiency of the path taken; (4) Normalized Error (NE) is the average distance between the agent's final position and the target, normalized by the initial distance to the target. These metrics collectively provide a holistic view of the model's navigation capabilities, evaluating not only its ability to reach the target but also the efficiency and accuracy of its path planning.

\subsection{Performance Comparison}
We assess OpenVLN in AirSim against three baselines: Fixed, CMA, and TravelUAV (SOTA). We analyze two aspects: (1) the effect of assistance levels on navigation, and (2) the effect of task difficulty on long-horizon performance.
As shown in Table \ref{tab:4-1:Ours_eval_results}, OpenVLN delivers consistent, significant gains over TravelUAV across all difficulties and assistance settings. Results are visualized in Fig. \ref{fig:eval}.

\noindent \textbf{Full Mode.}\; OpenVLN achieves substantial performance gains across all assistance levels. Under L1 assistance, we observe a 6.62-point reduction in NE from 132.59 to 125.97, coupled with improvements of 2.80\% in SR and 3.53\% in OSR. Under L2 assistance, the method delivers an 8.99-point NE reduction and 3.71\% SR improvement. Most notably, under the challenging L3 autonomous navigation scenario, OpenVLN achieves a substantial 12.93-point NE reduction and nearly doubles the SR from 1.39\% to 2.70\%. The consistent improvements across all assistance levels demonstrate our methods' robust adaptation to varying  conditions.

\noindent \textbf{Easy Mode.}\; In easier scenarios, OpenVLN maintains steady performance gains with more conservative improvements. Under L1 assistance, the method achieves a modest SR improvement from 14.11\% to 15.22\%. The L2 setting reveals the strength with an 8.54-point NE reduction and 4.40\% SR improvement from 12.90\% to 17.30\%. Under L3, the 7.47-point NE improvement from 100.37 to 92.90 demonstrates sustained performance even in autonomous navigation.The maintained effectiveness in easy evaluation validates the method's stability.

\noindent\textbf{Hard Mode.}\; The most significant performance gains emerge in challenging long-horizon scenarios. Under L1 assistance, OpenVLN achieves a 5.65-point NE reduction with a substantial 4.34\% SR improvement from 8.98\% to 13.32\%. L2 assistance produces even more pronounced gains with a 10.17-point NE reduction and 3.05\% SR improvement. Notably, under L3 assistance where the baseline completely fails with 0.00\% SR, OpenVLN maintains functionality with 0.47\% SR. The amplified performance advantage in hard scenarios confirms the method's superior handling of complex long-horizon planning challenges.

\subsection{Ablation Study}

To demonstrate the effect of reinforcement learning on model training, we conduct ablation study by adjusting diverse reward functions.We investigate the impact of different reward settings on navigation performance. As shown in Table \ref{tab:4-2:ablation-results}, the reward setting of our method(see at rows 1-3) is 5.0, derived from the dense reward mechanism described in Eq.\;(\ref{eq:dense-reward-machanism}).
 Others reward settings include 1.0, 3.0 and $\infty$ (see at rows 4-12), where 1.0 and 3.0 are manually set reward thresholds, and $\infty$ indicates no reward threshold limit.

\noindent \textbf{Full Mode.}\: With a reward threshold of 5.0, our method achieves optimal performance across all assistance levels. Under L1 assistance, it delivers 14.39\% SR, 28.03\% OSR, and 12.94\% SPL with an NE of 125.97. This significantly outperforms the 3.0 threshold (6.85\% / 14.26\% / 6.12\%), the 1.0 threshold (4.23\% / 9.39\% / 4.98\%), and the uncapped variant (1.31\% / 2.54\% / 1.05\%). Similar patterns emerge across L2 and L3 settings. Lower thresholds produce weak gradients that lead to underfitting. Conversely, removing the threshold generates excessively strong signals that destabilize training and reduce generalization.

\noindent \textbf{Easy Mode.}\: The reward scale of 5.0 demonstrates consistent superiority in less challenging scenarios. Under L1 assistance, our method achieves a SR of 15.22\% compared to 8.67\% with reward scale 3.0, representing a 6.55\% improvement. The OSR increases from 17.59\% to 30.64\%, while NE decreases from 103.95 to 87.96. Under L2 assistance, the performance advantage becomes more pronounced with SR improving from 6.79\% to 17.30\%, alongside a significant OSR enhancement from 16.68\% to 31.41\%. Even under autonomous L3 navigation, our method achieves 4.86\% SR while the 3.0 reward scale reaches only 3.42\%.  

\noindent \textbf{Hard Mode.}\: Performance advantages are most pronounced in challenging long-horizon scenarios. Under L1 assistance, our method achieves 13.32\% SR versus 6.04\% for reward scale 3.0, with NE improving from 214.80 to 175.54. Under L2, the SR gap widens further (10.24\% vs 6.36\%). Most notably, under demanding L3 conditions, our approach maintains 1.47\% SR while other scales degrade severely: 1.0 scale achieves only 0.86\% and the uncapped variant fails at 0.08\%.

Lower reward scales (1.0 and 3.0) result in underfitting due to insufficient learning that prevent effective policy optimization. Conversely, infinite reward scaling leads to overfitting and training instability, evidenced by poor performance across all metrics. The reward scaling mechanism critically affects the reinforcement learning dynamics, while setting reward as 5.0, is provides an effective balance between signal strength and stability, enabling robust policy updates under sparse feedback conditions while maintaining consistent performance across diverse navigation scenarios.

\section{CONCLUSIONS}
In this paper, we present an OpenVLN model, a data-efficient framework for Open-world aerial Vision-Language Navigation that unifies a value-model–guided, verifiable reward with reinforcement learning fine-tuning and a VLN-CE replanner for long-horizon control. The learned value model provides dense, per-waypoint, externally checkable feedback that improves credit assignment over long trajectories, while KL regularization stabilizes policy updates under scarce data. 

With only 25\% of training data, OpenVLN reduces NE on TravelUAV benchmarks and almost improves all the metrics performance across every assitant level, with performance gains of up to 4.34\% in Success Rate, 6.19\% in Oracle Success Rate, and 4.07\% in Success weighted by Path Length over baseline methods, evidencing stronger long-horizon robustness. Ablations confirm that the significance of value-guided dense rewards to balance exploration and stability Overall, verifiable value-shaped rewards with RL fine-tuning enable reliable waypoint synthesis under data scarcity and distribution shifts.


\addtolength{\textheight}{-12cm}   





\bibliography{cite.bib}

@inproceedings{andersonVision2018,
  title = {Vision-and-Language Navigation: {{Interpreting}} Visually-Grounded Navigation Instructions in Real Environments},
  shorttitle = {Vision-and-{{Language Navigation}}},
  booktitle = {2018 {{IEEE}}/{{CVF Conference}} on {{Computer Vision}} and {{Pattern Recognition}}},
  author = {Anderson, Peter and Wu, Qi and Teney, Damien and Bruce, Jake and Johnson, Mark and Sunderhauf, Niko and Reid, Ian and Gould, Stephen and Van Den Hengel, Anton},
  year = {2018},
  month = jun,
  pages = {3674--3683},
  publisher = {IEEE},
  address = {Salt Lake City, UT},
  doi = {10.1109/CVPR.2018.00387},
  urldate = {2025-09-14},
  isbn = {978-1-5386-6420-9},
  langid = {english}
}

@misc{chenConRFTReinforcedFinetuning2025,
  title = {{{ConRFT}}: A Reinforced Fine-Tuning Method for {{VLA}} Models via Consistency Policy},
  shorttitle = {{{ConRFT}}},
  author = {Chen, Yuhui and Tian, Shuai and Liu, Shugao and Zhou, Yingting and Li, Haoran and Zhao, Dongbin},
  year = {2025},
  month = apr,
  number = {arXiv:2502.05450},
  eprint = {2502.05450},
  primaryclass = {cs},
  publisher = {arXiv},
  doi = {10.48550/arXiv.2502.05450},
  urldate = {2025-09-14},
  archiveprefix = {arXiv},
  langid = {english}
}

@misc{lee2025citynavlargescaledatasetrealworld,
      title={CityNav: A Large-Scale Dataset for Real-World Aerial Navigation}, 
      author={Jungdae Lee and Taiki Miyanishi and Shuhei Kurita and Koya Sakamoto and Daichi Azuma and Yutaka Matsuo and Nakamasa Inoue},
      year={2025},
      eprint={2406.14240},
      archivePrefix={arXiv},
      primaryClass={cs.CV},
      url={https://arxiv.org/abs/2406.14240}, 
}

@misc{deepseek-aiDeepSeekR1IncentivizingReasoning2025,
  title = {{{DeepSeek-R1}}: {{Incentivizing}} Reasoning Capability in {{LLMs}} via Reinforcement Learning},
  shorttitle = {{{DeepSeek-R1}}},
  author = {{DeepSeek-AI} and Guo, Daya and Yang, Dejian and Zhang, Haowei and Song, Junxiao and Zhang, Ruoyu and Xu, Runxin and Zhu, Qihao and Ma, Shirong and Wang, Peiyi and Bi, Xiao and Zhang, Xiaokang and Yu, Xingkai and Wu, Yu and Wu, Z. F. and Gou, Zhibin and Shao, Zhihong and Li, Zhuoshu and Gao, Ziyi and Liu, Aixin and Xue, Bing and Wang, Bingxuan and Wu, Bochao and Feng, Bei and Lu, Chengda and Zhao, Chenggang and Deng, Chengqi and Zhang, Chenyu and Ruan, Chong and Dai, Damai and Chen, Deli and Ji, Dongjie and Li, Erhang and Lin, Fangyun and Dai, Fucong and Luo, Fuli and Hao, Guangbo and Chen, Guanting and Li, Guowei and Zhang, H. and Bao, Han and Xu, Hanwei and Wang, Haocheng and Ding, Honghui and Xin, Huajian and Gao, Huazuo and Qu, Hui and Li, Hui and Guo, Jianzhong and Li, Jiashi and Wang, Jiawei and Chen, Jingchang and Yuan, Jingyang and Qiu, Junjie and Li, Junlong and Cai, J. L. and Ni, Jiaqi and Liang, Jian and Chen, Jin and Dong, Kai and Hu, Kai and Gao, Kaige and Guan, Kang and Huang, Kexin and Yu, Kuai and Wang, Lean and Zhang, Lecong and Zhao, Liang and Wang, Litong and Zhang, Liyue and Xu, Lei and Xia, Leyi and Zhang, Mingchuan and Zhang, Minghua and Tang, Minghui and Li, Meng and Wang, Miaojun and Li, Mingming and Tian, Ning and Huang, Panpan and Zhang, Peng and Wang, Qiancheng and Chen, Qinyu and Du, Qiushi and Ge, Ruiqi and Zhang, Ruisong and Pan, Ruizhe and Wang, Runji and Chen, R. J. and Jin, R. L. and Chen, Ruyi and Lu, Shanghao and Zhou, Shangyan and Chen, Shanhuang and Ye, Shengfeng and Wang, Shiyu and Yu, Shuiping and Zhou, Shunfeng and Pan, Shuting and Li, S. S. and Zhou, Shuang and Wu, Shaoqing and Ye, Shengfeng and Yun, Tao and Pei, Tian and Sun, Tianyu and Wang, T. and Zeng, Wangding and Zhao, Wanjia and Liu, Wen and Liang, Wenfeng and Gao, Wenjun and Yu, Wenqin and Zhang, Wentao and Xiao, W. L. and An, Wei and Liu, Xiaodong and Wang, Xiaohan and Chen, Xiaokang and Nie, Xiaotao and Cheng, Xin and Liu, Xin and Xie, Xin and Liu, Xingchao and Yang, Xinyu and Li, Xinyuan and Su, Xuecheng and Lin, Xuheng and Li, X. Q. and Jin, Xiangyue and Shen, Xiaojin and Chen, Xiaosha and Sun, Xiaowen and Wang, Xiaoxiang and Song, Xinnan and Zhou, Xinyi and Wang, Xianzu and Shan, Xinxia and Li, Y. K. and Wang, Y. Q. and Wei, Y. X. and Zhang, Yang and Xu, Yanhong and Li, Yao and Zhao, Yao and Sun, Yaofeng and Wang, Yaohui and Yu, Yi and Zhang, Yichao and Shi, Yifan and Xiong, Yiliang and He, Ying and Piao, Yishi and Wang, Yisong and Tan, Yixuan and Ma, Yiyang and Liu, Yiyuan and Guo, Yongqiang and Ou, Yuan and Wang, Yuduan and Gong, Yue and Zou, Yuheng and He, Yujia and Xiong, Yunfan and Luo, Yuxiang and You, Yuxiang and Liu, Yuxuan and Zhou, Yuyang and Zhu, Y. X. and Xu, Yanhong and Huang, Yanping and Li, Yaohui and Zheng, Yi and Zhu, Yuchen and Ma, Yunxian and Tang, Ying and Zha, Yukun and Yan, Yuting and Ren, Z. Z. and Ren, Zehui and Sha, Zhangli and Fu, Zhe and Xu, Zhean and Xie, Zhenda and Zhang, Zhengyan and Hao, Zhewen and Ma, Zhicheng and Yan, Zhigang and Wu, Zhiyu and Gu, Zihui and Zhu, Zijia and Liu, Zijun and Li, Zilin and Xie, Ziwei and Song, Ziyang and Pan, Zizheng and Huang, Zhen and Xu, Zhipeng and Zhang, Zhongyu and Zhang, Zhen},
  year = {2025},
  month = jan,
  number = {arXiv:2501.12948},
  eprint = {2501.12948},
  primaryclass = {cs},
  publisher = {arXiv},
  doi = {10.48550/arXiv.2501.12948},
  urldate = {2025-09-14},
  archiveprefix = {arXiv}
}

@misc{gaoOpenFly2025,
  title = {{{OpenFly}}: A Comprehensive Platform for Aerial Vision-Language Navigation},
  shorttitle = {{{OpenFly}}},
  author = {Gao, Yunpeng and Li, Chenhui and You, Zhongrui and Liu, Junli and Li, Zhen and Chen, Pengan and Chen, Qizhi and Tang, Zhonghan and Wang, Liansheng and Yang, Penghui and Tang, Yiwen and Tang, Yuhang and Liang, Shuai and Zhu, Songyi and Xiong, Ziqin and Su, Yifei and Ye, Xinyi and Li, Jianan and Ding, Yan and Wang, Dong and Wang, Zhigang and Zhao, Bin and Li, Xuelong},
  year = {2025},
  month = jul,
  number = {arXiv:2502.18041},
  eprint = {2502.18041},
  primaryclass = {cs},
  publisher = {arXiv},
  doi = {10.48550/arXiv.2502.18041},
  urldate = {2025-09-14},
  archiveprefix = {arXiv},
  langid = {english}
}

@inproceedings{guptaCognitiveMappingPlanning2017,
  title = {Cognitive Mapping and Planning for Visual Navigation},
  booktitle = {Proceedings of the {{IEEE Conference}} on {{Computer Vision}} and {{Pattern Recognition}}},
  author = {Gupta, Saurabh and Davidson, James and Levine, Sergey and Sukthankar, Rahul and Malik, Jitendra},
  year = {2017},
  pages = {2616--2625},
  urldate = {2025-09-14},
  langid = {english}
}

@inproceedings{ichterCanNotSay2023,
  title = {Do as {{I}} Can, Not as {{I}} Say: Grounding Language in Robotic Affordances},
  shorttitle = {Do {{As I}} Can, Not {{As I}} Say},
  booktitle = {Proceedings of the 6th {{Conference}} on {{Robot Learning}}},
  author = {Ichter, Brian and Brohan, Anthony and Chebotar, Yevgen and Finn, Chelsea and Hausman, Karol and Herzog, Alexander and Ho, Daniel and Ibarz, Julian and Irpan, Alex and Jang, Eric and Julian, Ryan and Kalashnikov, Dmitry and Levine, Sergey and Lu, Yao and Parada, Carolina and Rao, Kanishka and Sermanet, Pierre and Toshev, Alexander T. and Vanhoucke, Vincent and Xia, Fei and Xiao, Ted and Xu, Peng and Yan, Mengyuan and Brown, Noah and Ahn, Michael and Cortes, Omar and Sievers, Nicolas and Tan, Clayton and Xu, Sichun and Reyes, Diego and Rettinghouse, Jarek and Quiambao, Jornell and Pastor, Peter and Luu, Linda and Lee, Kuang-Huei and Kuang, Yuheng and Jesmonth, Sally and Joshi, Nikhil J. and Jeffrey, Kyle and Ruano, Rosario Jauregui and Hsu, Jasmine and Gopalakrishnan, Keerthana and David, Byron and Zeng, Andy and Fu, Chuyuan Kelly},
  year = {2023},
  month = mar,
  pages = {287--318},
  publisher = {PMLR},
  issn = {2640-3498},
  urldate = {2025-09-14},
  langid = {english}
}

@inproceedings{krantz2020,
  title = {Beyond the Nav-Graph: Vision-and-Language Navigation in Continuous Environments},
  shorttitle = {Beyond the Nav-Graph},
  booktitle = {Computer {{Vision}} -- {{ECCV}} 2020},
  author = {Krantz, Jacob and Wijmans, Erik and Majumdar, Arjun and Batra, Dhruv and Lee, Stefan},
  editor = {Vedaldi, Andrea and Bischof, Horst and Brox, Thomas and Frahm, Jan-Michael},
  year = {2020},
  pages = {104--120},
  publisher = {Springer International Publishing},
  address = {Cham},
  doi = {10.1007/978-3-030-58604-1_7},
  isbn = {978-3-030-58604-1},
  langid = {english}
}

@inproceedings{krantzSim2simTransferVisionandlanguage2022,
  title = {Sim-2-Sim Transfer for Vision-and-Language Navigation in Continuous Environments},
  booktitle = {Computer {{Vision}} -- {{ECCV}} 2022},
  author = {Krantz, Jacob and Lee, Stefan},
  year = {2022},
  pages = {588--603},
  publisher = {Springer, Cham},
  issn = {1611-3349},
  doi = {10.1007/978-3-031-19842-7_34},
  urldate = {2025-09-14},
  isbn = {978-3-031-19842-7},
  langid = {english}
}

@misc{dorbala2022clipnavusingclipzeroshot,
      title={CLIP-Nav: Using CLIP for Zero-Shot Vision-and-Language Navigation}, 
      author={Vishnu Sashank Dorbala and Gunnar Sigurdsson and Robinson Piramuthu and Jesse Thomason and Gaurav S. Sukhatme},
      year={2022},
      eprint={2211.16649},
      archivePrefix={arXiv},
      primaryClass={cs.CV},
      url={https://arxiv.org/abs/2211.16649}, 
}

@misc{kulkarniDeepSuccessorReinforcement2016,
  title = {Deep Successor Reinforcement Learning},
  author = {Kulkarni, Tejas D. and Saeedi, Ardavan and Gautam, Simanta and Gershman, Samuel J.},
  year = {2016},
  month = jun,
  number = {arXiv:1606.02396},
  eprint = {1606.02396},
  primaryclass = {stat},
  publisher = {arXiv},
  doi = {10.48550/arXiv.1606.02396},
  urldate = {2025-09-14},
  archiveprefix = {arXiv},
  langid = {english}
}

@inproceedings{liuAerialVLNVisionandlanguageNavigation2023,
  title = {{{AerialVLN}}: Vision-and-Language Navigation for {{UAVs}}},
  shorttitle = {{{AerialVLN}}},
  booktitle = {Proceedings of the {{IEEE}}/{{CVF International Conference}} on {{Computer Vision}}},
  author = {Liu, Shubo and Zhang, Hongsheng and Qi, Yuankai and Wang, Peng and Zhang, Yanning and Wu, Qi},
  year = {2023},
  pages = {15384--15394},
  urldate = {2025-09-14},
  langid = {english}
}

@inproceedings{liuGroundingDINOMarrying2025,
  title = {Grounding {{DINO}}: Marrying {{DINO}} with~Grounded Pre-Training for~Open-Set Object Detection},
  shorttitle = {Grounding {{DINO}}},
  booktitle = {Computer {{Vision}} -- {{ECCV}} 2024},
  author = {Liu, Shilong and Zeng, Zhaoyang and Ren, Tianhe and Li, Feng and Zhang, Hao and Yang, Jie and Jiang, Qing and Li, Chunyuan and Yang, Jianwei and Su, Hang and Zhu, Jun and Zhang, Lei},
  editor = {Leonardis, Ale{\v s} and Ricci, Elisa and Roth, Stefan and Russakovsky, Olga and Sattler, Torsten and Varol, G{\"u}l},
  year = {2025},
  pages = {38--55},
  publisher = {Springer Nature Switzerland},
  address = {Cham},
  doi = {10.1007/978-3-031-72970-6_3},
  isbn = {978-3-031-72970-6},
  langid = {english}
}

@misc{liuRDT1BDiffusionFoundation2025,
  title = {{{RDT-1B}}: A Diffusion Foundation Model for Bimanual Manipulation},
  shorttitle = {Rdt-1b},
  author = {Liu, Songming and Wu, Lingxuan and Li, Bangguo and Tan, Hengkai and Chen, Huayu and Wang, Zhengyi and Xu, Ke and Su, Hang and Zhu, Jun},
  year = {2025},
  month = mar,
  number = {arXiv:2410.07864},
  eprint = {2410.07864},
  primaryclass = {cs},
  publisher = {arXiv},
  doi = {10.48550/arXiv.2410.07864},
  urldate = {2025-09-14},
  archiveprefix = {arXiv},
  langid = {english}
}

@misc{liuVisualRFTVisualReinforcement2025,
  title = {Visual-{{RFT}}: {{Visual Reinforcement Fine-Tuning}}},
  shorttitle = {Visual-{{RFT}}},
  author = {Liu, Ziyu and Sun, Zeyi and Zang, Yuhang and Dong, Xiaoyi and Cao, Yuhang and Duan, Haodong and Lin, Dahua and Wang, Jiaqi},
  year = {2025},
  month = mar,
  number = {arXiv:2503.01785},
  eprint = {2503.01785},
  primaryclass = {cs},
  publisher = {arXiv},
  doi = {10.48550/arXiv.2503.01785},
  urldate = {2025-08-29},
  archiveprefix = {arXiv},
  langid = {english}
}

@inproceedings{LLaMAVIDImageWorth,
  title={Llama-vid: An image is worth 2 tokens in large language models},
  author={Li, Yanwei and Wang, Chengyao and Jia, Jiaya},
  booktitle={European Conference on Computer Vision},
  pages={323--340},
  year={2024},
  organization={Springer}
}

@article{NavilaLeggedRobot,
  title={Navila: Legged robot vision-language-action model for navigation},
  author={Cheng, An-Chieh and Ji, Yandong and Yang, Zhaojing and Gongye, Zaitian and Zou, Xueyan and Kautz, Jan and B{\i}y{\i}k, Erdem and Yin, Hongxu and Liu, Sifei and Wang, Xiaolong},
  journal={arXiv preprint arXiv:2412.04453},
  year={2024}
}

@misc{qiREVERIERemoteEmbodied2020,
  title = {{{REVERIE}}: Remote Embodied Visual Referring Expression in Real Indoor Environments},
  shorttitle = {Reverie},
  author = {Qi, Yuankai and Wu, Qi and Anderson, Peter and Wang, Xin and Wang, William Yang and Shen, Chunhua and van den Hengel, Anton},
  year = {2020},
  month = jan,
  number = {arXiv:1904.10151},
  eprint = {1904.10151},
  primaryclass = {cs},
  publisher = {arXiv},
  doi = {10.48550/arXiv.1904.10151},
  urldate = {2025-09-14},
  archiveprefix = {arXiv},
  langid = {english}
}

@misc{shuRFTFReinforcementFinetuning2025,
  title = {{{RFTF}}: {{Reinforcement Fine-tuning}} for {{Embodied Agents}} with {{Temporal Feedback}}},
  shorttitle = {{{RFTF}}},
  author = {Shu, Junyang and Lin, Zhiwei and Wang, Yongtao},
  year = {2025},
  month = may,
  number = {arXiv:2505.19767},
  eprint = {2505.19767},
  primaryclass = {cs},
  publisher = {arXiv},
  doi = {10.48550/arXiv.2505.19767},
  urldate = {2025-08-29},
  archiveprefix = {arXiv},
  langid = {english}
}

@article{ouyang2022training,
  title={Training language models to follow instructions with human feedback},
  author={Ouyang, Long and Wu, Jeffrey and Jiang, Xu and Almeida, Diogo and Wainwright, Carroll and Mishkin, Pamela and Zhang, Chong and Agarwal, Sandhini and Slama, Katarina and Ray, Alex and others},
  journal={Advances in neural information processing systems},
  volume={35},
  pages={27730--27744},
  year={2022}
}

@misc{saxena2025uavvlnendtoendvisionlanguage,
      title={{{UAV-VLN}}:  End-to-End Vision Language guided Navigation for UAVs}, 
      author={Pranav Saxena and Nishant Raghuvanshi and Neena Goveas},
      year={2025},
      eprint={2504.21432},
      archivePrefix={arXiv},
      primaryClass={cs.RO},
      url={https://arxiv.org/abs/2504.21432}, 
}

@misc{wangRealisticUAVVisionlanguage2024,
  title = {Towards Realistic {{UAV}} Vision-Language Navigation: {{Platform}}, Benchmark, and Methodology},
  shorttitle = {Towards {{Realistic UAV Vision-Language Navigation}}},
  author = {Wang, Xiangyu and Yang, Donglin and Wang, Ziqin and Kwan, Hohin and Chen, Jinyu and Wu, Wenjun and Li, Hongsheng and {Yue Liao} and Liu, Si},
  year = {2024},
  month = oct,
  number = {arXiv:2410.07087},
  eprint = {2410.07087},
  primaryclass = {cs},
  publisher = {arXiv},
  doi = {10.48550/arXiv.2410.07087},
  urldate = {2025-06-20},
  archiveprefix = {arXiv},
  langid = {american}
}

@misc{wangUAVflowColosseoRealworld2025,
  title = {{{UAV-flow}} Colosseo: A Real-World Benchmark for Flying-on-a-Word {{UAV}} Imitation Learning},
  shorttitle = {{{UAV-flow}} Colosseo},
  author = {Wang, Xiangyu and Yang, Donglin and Liao, Yue and Zheng, Wenhao and Wu, Wenjun and Dai, Bin and Li, Hongsheng and Liu, Si},
  year = {2025},
  month = may,
  number = {arXiv:2505.15725},
  eprint = {2505.15725},
  primaryclass = {cs},
  publisher = {arXiv},
  doi = {10.48550/arXiv.2505.15725},
  urldate = {2025-06-20},
  archiveprefix = {arXiv},
  langid = {english}
}

@article{yueSafeVLNCollisionAvoidance2024,
  title = {Safe-{{VLN}}: Collision Avoidance for Vision-and-Language Navigation of Autonomous Robots Operating in Continuous Environments},
  shorttitle = {Safe-{{VLN}}},
  author = {Yue, Lu and Zhou, Dongliang and Xie, Liang and Zhang, Feitian and Yan, Ye and Yin, Erwei},
  year = {2024},
  month = jun,
  journal = {IEEE Robotics and Automation Letters},
  volume = {9},
  number = {6},
  pages = {4918--4925},
  issn = {2377-3766},
  doi = {10.1109/LRA.2024.3387171},
  urldate = {2025-06-20},
  langid = {english}
}

@misc{zangInternLMXComposer25rewardSimpleEffective2025,
  title = {{{InternLM-XComposer2}}.5-Reward: A Simple yet Effective Multi-Modal Reward Model},
  shorttitle = {{{InternLM-XComposer2}}.5-Reward},
  author = {Zang, Yuhang and Dong, Xiaoyi and Zhang, Pan and Cao, Yuhang and Liu, Ziyu and Ding, Shengyuan and Wu, Shenxi and Ma, Yubo and Duan, Haodong and Zhang, Wenwei and Chen, Kai and Lin, Dahua and Wang, Jiaqi},
  year = {2025},
  month = may,
  number = {arXiv:2501.12368},
  eprint = {2501.12368},
  primaryclass = {cs},
  publisher = {arXiv},
  doi = {10.48550/arXiv.2501.12368},
  urldate = {2025-09-14},
  archiveprefix = {arXiv},
  langid = {english}
}

@misc{zhangNaVidVideobasedVLM2024,
  title = {{{NaVid}}: Video-Based {{VLM}} Plans the next Step for Vision-and-Language Navigation},
  shorttitle = {{{NaVid}}},
  author = {Zhang, Jiazhao and Wang, Kunyu and Xu, Rongtao and Zhou, Gengze and Hong, Yicong and Fang, Xiaomeng and Wu, Qi and Zhang, Zhizheng and Wang, He},
  year = {2024},
  month = jun,
  number = {arXiv:2402.15852},
  eprint = {2402.15852},
  primaryclass = {cs},
  publisher = {arXiv},
  doi = {10.48550/arXiv.2402.15852},
  urldate = {2025-09-14},
  archiveprefix = {arXiv},
  langid = {english}
}

@inproceedings{zhouEGOswarmFullyAutonomous2021,
  title = {{{EGO-swarm}}: A Fully Autonomous and Decentralized Quadrotor Swarm System in Cluttered Environments},
  shorttitle = {{{EGO-swarm}}},
  booktitle = {2021 {{IEEE International Conference}} on {{Robotics}} and {{Automation}} ({{ICRA}})},
  author = {Zhou, Xin and Zhu, Jiangchao and Zhou, Hongyu and Xu, Chao and Gao, Fei},
  year = {2021},
  month = may,
  pages = {4101--4107},
  issn = {2577-087X},
  doi = {10.1109/ICRA48506.2021.9561902},
  urldate = {2025-09-14},
  langid = {english}
}

@article{zhouNavGPTExplicitReasoning2024,
  title = {{{NavGPT}}: Explicit Reasoning in Vision-and-Language Navigation with Large Language Models},
  shorttitle = {{{NavGPT}}},
  author = {Zhou, Gengze and Hong, Yicong and Wu, Qi},
  year = {2024},
  month = mar,
  journal = {Proceedings of the AAAI Conference on Artificial Intelligence},
  volume = {38},
  number = {7},
  pages = {7641--7649},
  issn = {2374-3468},
  doi = {10.1609/aaai.v38i7.28597},
  urldate = {2025-09-14},
  copyright = {Copyright (c) 2024 Association for the Advancement of Artificial Intelligence},
  langid = {english}
}

@inproceedings{zhuSOONScenarioOriented2021a,
  title = {{{SOON}}: Scenario Oriented Object Navigation with Graph-Based Exploration},
  shorttitle = {Soon},
  booktitle = {Proceedings of the {{IEEE}}/{{CVF Conference}} on {{Computer Vision}} and {{Pattern Recognition}}},
  author = {Zhu, Fengda and Liang, Xiwen and Zhu, Yi and Yu, Qizhi and Chang, Xiaojun and Liang, Xiaodan},
  year = {2021},
  pages = {12689--12699},
  urldate = {2025-09-14},
  langid = {english}
}

@inproceedings{zhuTargetdrivenVisualNavigation2017,
  title = {Target-Driven Visual Navigation in Indoor Scenes Using Deep Reinforcement Learning},
  booktitle = {2017 {{IEEE International Conference}} on {{Robotics}} and {{Automation}} ({{ICRA}})},
  author = {Zhu, Yuke and Mottaghi, Roozbeh and Kolve, Eric and Lim, Joseph J. and Gupta, Abhinav and {Fei-Fei}, Li and Farhadi, Ali},
  year = {2017},
  month = may,
  pages = {3357--3364},
  doi = {10.1109/ICRA.2017.7989381},
  urldate = {2025-09-14},
  langid = {english}
}

\end{document}